\documentclass[11pt,a4paper]{article}
\usepackage[hyperref]{acl2020}
\usepackage{times}
\usepackage{latexsym}

\usepackage{microtype}
\usepackage{times}
\usepackage{latexsym}
\usepackage{graphicx}
\usepackage{xspace}
\usepackage{multirow}
\usepackage{xcolor}
\usepackage{url}
\usepackage{arydshln}
\usepackage{amssymb}
\usepackage{pifont}
\usepackage{pbox}

\newcommand{\xmark}{\ding{55}}%
\aclfinalcopy 
\def\aclpaperid{3195}

\newcommand\perspective{\textsc{perspective}\xspace}
\newcommand\caperspective{\textsc{ca-conc-perspective}\xspace}
\newcommand\convai{\textsc{cwtk}\xspace}
\newcommand\largedataset{\textsc{cat-large}\xspace}
\newcommand\smalldataset{\textsc{cat-small}\xspace}

\newcommand\mc{\textsc{mc}\xspace}
\newcommand\oc{\textsc{n}\xspace}
\newcommand\ic{\textsc{c}\xspace}
\newcommand\goc{\textsc{gn}\xspace}
\newcommand\gic{\textsc{gc}\xspace}
\newcommand\civilcomments{\textsc{cctk}\xspace}
\newcommand\offenseval{\textsc{OffensEval}\xspace}

\newcommand\auc{\textsc{auc}\xspace}
\newcommand\roc{\textsc{roc}\xspace}
\newcommand\lstm{\textsc{lstm}\xspace}
\newcommand\bilstm{\textsc{bilstm}\xspace}
\newcommand\cabilstm{\textsc{ca-bilstm-bilstm}\xspace}

\newcommand\ffnn{\textsc{ffnn}\xspace}
\newcommand\bert{\textsc{bert}\xspace}
\newcommand\cabert{\textsc{ca-bilstm-bert}\xspace}
\newcommand\casepbert{\textsc{ca-sep-bert}\xspace}

\newcommand\bertbase{\textsc{bert-base}\xspace}
\newcommand\bertcctk{\textsc{bert-cctk}\xspace}
\newcommand\cabertcctk{\textsc{ca-conc-bert-cctk}\xspace}
\newcommand\rnn{\textsc{rnn}\xspace}
\newcommand\cnn{\textsc{cnn}\xspace}
\newcommand\lda{\textsc{lda}\xspace}
\newcommand\bleu{\textsc{bleu}\xspace}
\newcommand\rqone{\textsc{rq1}\xspace}
\newcommand\rqtwo{\textsc{rq2}\xspace}

\title{Toxicity Detection: Does Context Really Matter?}

\author{
	John Pavlopoulos$^\dagger$, 
	Jeffrey Sorensen$^\ddagger$,  
	Lucas Dixon$^\ddagger$, 
	Nithum Thain$^\ddagger$, 
	Ion Androutsopoulos$^\dagger$ \\
	$^\dagger$ {Department of Informatics, Athens University of Economic and Business, Greece}\\
	{\tt {annis,ion}@aueb.gr} \\
	$^\ddagger$ Google\\
	{\tt {sorenj,ldixon,nthain}@google.com}
}

\date{}

\begin{document}

\maketitle

\begin{abstract}
Moderation is crucial to promoting healthy online discussions. Although several `toxicity' detection datasets and models have been published, most of them ignore the context of the posts, implicitly assuming that comments may be judged independently.
We investigate this assumption by focusing on two questions: (a) does context affect the  human judgement, and (b) does conditioning on context improve performance of toxicity detection systems?
We experiment with Wikipedia conversations, limiting the notion of context to the previous post in the thread and the discussion title. We find that context can both amplify or mitigate the perceived toxicity of posts. Moreover, a small but significant subset of manually labeled posts (5\% in one of our experiments) 
end up having the opposite toxicity labels if the annotators are not provided with context.
Surprisingly, we also find no evidence that context actually improves the performance of toxicity classifiers, having tried a range of classifiers and mechanisms to make them context aware.
This points to the need for larger datasets of  comments annotated in context.
We make our code and data publicly available.
\end{abstract}

\section{Introduction}

\begin{table}
\small
    \begin{center}
    \begin{tabular}{|p{1cm}|p{5.8cm}| }
    \hline

    \textsc{Parent} & All of his arguements are nail perfect, you're 
    inherently stupid. The lead will be changed. \\ 
    \textsc{Target} & \textbf{Great argument!} \\
    \hline\hline

    \textsc{Parent} & Really?  It's schmucks like you (and Bush) who turn the world into the shithole it is today! \\ 
    \textsc{Target} & \textbf{I'd be interested in the reasoning for that comment, personally.   (bounties)} \\
    \hline\hline

    \textsc{Parent} &  Indeed.  Hitler was also strongly anti-pornography [\dots] it sure looks like Hitler is a hot potato that nobody wants to be stuck with. \\ 
    \textsc{Target} & \textbf{Well I guess they won't approve the slogan ``Hitler hated porn''.} \\
    \hline\hline

    \textsc{Parent} &  ?? When did I attack you? I definitely will present this to the arbcom, you should mind WP:CIVIL when participating in discussions in Wikipedia. \\ 
    \textsc{Target} & \textbf{I blame you for my alcoholism add that too} \\
    \hline

    \end{tabular}
    \caption{Comments that are not easily labeled for toxicity without the `parent' (previous) comment. The `target' comment is the one being labeled.} 
    \vspace*{-4mm}
    \label{tab:first_examples}
    \end{center}
\end{table}

Systems that detect abusive language are used to promote healthy conversations online and protect minority voices \cite{Hosseini2017}. Apart from a growing volume of press articles concerning
toxicity
online,\footnote{Following the work of \citet{Wulczyn2017} and \citet{10.1145/3308560.3317593}, \emph{toxicity} is defined as ``a rude, disrespectful, or unreasonable comment that is likely to make you leave a discussion'' \cite{Wulczyn2017}.}
there is increased research interest on detecting 
abusive and other unwelcome  
comments labeled `toxic' by moderators, both for English and other languages.\footnote{For English, see for example \textsc{trac} \cite{trac2018report}, \offenseval \cite{Offenseval}, or the recent Workshops on Abusive Language Online (\url{https://goo.gl/9HmSzc}). For other languages, see for example the German \textsc{GermEval} (\url{https://goo.gl/uZEerk}).}
However, the vast majority of current datasets do not include the preceding comments in a conversation and such context was not shown to the annotators who provided the gold toxicity labels. 
Consequently, systems trained on these datasets ignore the conversational context. For example, a comment like ``nope, I don't think so'' may not be judged as rude or inflammatory by such a system, but the system's score would probably be higher if the system could also consider the previous (also called \emph{parent}) comment ``might it be that I am sincere?''.
Table~\ref{tab:first_examples} shows additional examples of comments that are not easily judged for toxicity without the parent comment. Interestingly, even basic statistics on how often context affects the perceived toxicity of online posts have not been 
published. Hence, in this paper we focus on the following two foundational research questions:

\begin{itemize} 
     \item \rqone: \textit{How often does context affect the  toxicity of posts as perceived by humans in online conversations? And how often does context amplify or mitigate the perceived toxicity?}
    \item \rqtwo: \textit{Does context actually improve the performance of toxicity classifiers, when they are made context-aware? And how can toxicity classifiers be made context-aware?}
\end{itemize}

To investigate these questions we created and make publicly available two new toxicity datasets that include context, which are based on discussions in Wikipedia Talk Pages \cite{Hua2018}. The first one is a small dataset of 250 comments, created in an \textsc{AB} test fashion, where two different groups of annotators (crowd-workers) were employed. One group annotated the comments without context, while the other group was given the same comments, this time along with the parent comment and the title of the thread as context. We used this dataset to show that the perceived toxicity of a significant subset of posts (5.2\% in our experiment) changes when context is (or is not) provided. We conclude that a small but significant subset of manually labeled posts end up having wrong toxicity labels if the annotators are not provided with context. We also found that context can both amplify (approximately 3.6\% of comments in our experiment) and mitigate (approx.\ 1.6\%) the perceived toxicity. Examples of comments that were differently labeled with and without context are shown in Table~\ref{tab:annotated_examples}.

\begin{table}
\small
    \begin{center}
    \begin{tabular}{|p{1cm}|p{5.9cm}|} 
    \multicolumn{2}{c}{\textsc{Comment with toxicity AMPLIFIED in context}} \\\hline
    \textsc{Parent} & But what if the user is a lesbian? Then what? \\ 
    \textsc{Target} & \textbf{``Pigs Are People Too".\ ``Avant-garde a clue''}
    \\ \hline
    \multicolumn{2}{c}{\textsc{Comment with toxicity MITIGATED in context}} \\\hline
    \textsc{Parent} & Hmmm.  The flame on top of the gay pride emblem can probably be interpreted in a manner that I did not consider.  Perhaps one icon on each end using?\\ 
    \textsc{Target} & \textbf{Hi Gadget, interpreted in what manner? Flaming gays? Or Burn a gay?} \\\hline
    \end{tabular}
    \caption{Examples of comments that the annotators labeled differently when the previous (parent) comment was (or not) provided. In the top example, the target comment (the one being annotated) was labeled as toxic only when context was given. In the bottom example, the target comment was considered toxic only without its parent comment.} 
    \vspace*{-4mm}
    \label{tab:annotated_examples}
    \end{center}
\end{table}

To investigate the second question, concerning the effect of context on the performance of toxicity classifiers, we created a larger dataset of 20k comments; 10k comments were annotated out of context, 10k in context. This time we did not require the \emph{same} comments to be annotated with and without context, which allowed us to crowd-source the collection of a larger set of annotations.
These two new subsets were used to train several toxicity detection classifiers, both context-aware and context-unaware, which were evaluated on held out comments that we always annotated in context (based on the assumption that in-context labels are more reliable).
Surprisingly, we found no evidence that context actually improves the performance of toxicity classifiers. We tried a range of classifiers and mechanisms to make them context aware, and having also considered the effect of using gold labels obtained out of context or by showing context to the annotators. 
This finding is likely related to the small number of
context-sensitive comments. 
In turn this suggests that an important direction for further research is how to efficiently annotate 
larger corpora of comments in context. We make our code and data publicly available.\footnote{\url{https://github.com/ipavlopoulos/context_toxicity}}

\section{Related Work}
\label{sec:related_work}

\begin{table*}
    \small
    \begin{center}
    \begin{tabular}{ |c|c|c|c|c||c|c| }
    \hline
    \textbf{Dataset Name} & \textbf{Source} & \textbf{Size} & \textbf{Type} & \textbf{Lang.} & $C_{a}$ & $C_{t}$ \\\hline\hline
    \civilcomments & Civil Comments Toxicity Kaggle & 2M & Toxicity sub-types & EN & \xmark & - \\\hline
    \convai & Wikipedia Toxicity Kaggle & 223,549 & Toxicity sub-types & EN & \xmark & - \\ \hline
    \newcite{davidson2017automated} & Twitter & 24,783 & Hate/Offense & EN & \xmark & - \\\hline
    \newcite{OLID} & Twitter & 14,100 & Offense & EN & \xmark & - \\\hline
    \newcite{Waseem2016} & Twitter & 1,607 & Sexism/Racism & EN & \xmark & - \\ \hline
    \newcite{Gao2018} & Fox News & 1,528 & Hate & EN & \checkmark & {\color{red}\tiny Title} \\\hline
    \newcite{wiegand2018overview} & Twitter & 8541 & Insult/Abuse/Profanity & DE & \xmark & - \\\hline
    \newcite{ross2016measuring} & Twitter & 470 & Hate & DE & \xmark & - \\ \hline
    \newcite{Pavlopoulos2017a} & \url{Gazzetta.gr} & 1,6M & Rejection & EL & \checkmark & - \\ \hline
    \newcite{mubarak2017} & \url{Aljazeera.net} & 31,633 & Obscene/Offense & AR & \checkmark & {\color{red}\tiny Title} \\\hline 
    \end{tabular}
    \caption{Publicly available datasets for toxicity detection. The Size column shows the number of comments. Column $C_{a}$ shows if annotation was context-aware or not. Column $C_{t}$ shows the type of context provided. \newcite{Pavlopoulos2017a} used professional moderator decisions, which were context-aware, but context is not included in their dataset. 
    The datasets of \newcite{Gao2018} and \newcite{mubarak2017} include context-aware labels, but provide only the titles of the news articles being discussed.
    }
    \vspace*{-4mm}
    \label{tab:datasets}
    \end{center}
\end{table*}

Toxicity detection has attracted a lot of attention in 
recent years \cite{Nobata2016,Pavlopoulos2017b,Park2017,Wulczyn2017}. 
Here we use the term `toxic' as an umbrella term, but we note that the literature uses several terms for different kinds of toxic language or related phenomena: `offensive' \cite{OLID}, `abusive' \cite{Pavlopoulos2017a}, `hateful' \cite{Djuric2015,malmasi2017detecting,elsherief2018hate,gamback2017using,zhang2018detecting}, etc. There are also taxonomies for these phenomena based on their directness (e.g., whether the abuse was unambiguously implied/denoted or not), and their target (e.g., whether it was a general comment or targeting an individual/group) \cite{waseem2017understanding}. Other hierarchical taxonomies have also been defined \cite{OLID}. While most previous work does not address toxicity in general, instead addressing particular subtypes, toxicity and its subtypes are strongly related, with systems trained to detect toxicity being effective also at subtypes, such as hateful language \cite{VanAken2018}. 
As is customary in natural language processing, we focus on aggregate results when hoping to answer our research questions, and leave largely unanswered the related epistemological questions when this does not preclude 
using classifiers in real-world applications.

Table~\ref{tab:datasets} lists all currently available public datasets for the various forms of toxic language that we are aware of. The two last columns show that no existing English dataset provides both context (e.g.,  parent comment) and context-aware annotations (annotations provided by humans who also considered the parent comment).

Both small and large toxicity datasets have been developed, but approximately half of them contain tweets, which makes reusing the data difficult, because abusive tweets are often removed by the platform. Moreover, the textual content is not available under a license that allows its storage outside the platform. 
The hateful language detection dataset of \newcite{Waseem2016}, for example, contains 1,607 sexism and racism annotations for \textsc{id}s of English tweets. A larger dataset was published by \newcite{davidson2017automated}, containing approx.\ 25k annotations for tweet-\textsc{id}s, collected using a lexicon of hateful terms.
Research on forms of abusive language detection is mainly focused on English (6 out of 10 datasets), but datasets in other languages also exist, such as Greek \cite{Pavlopoulos2017a}, Arabic \cite{mubarak2017}, and German \cite{ross2016measuring,wiegand2018overview}. 

A common characteristic of most of the datasets listed in Table~\ref{tab:datasets} is that, during annotation, the human workers were not provided with, nor instructed to review, the context of the target text. Context such as the preceding comments in the thread, or the title of the article being discussed, or the discussion topic. A notable exception is the work of \newcite{Gao2018}, who annotated hateful comments under Fox News articles by also considering the title of the news article and the preceding comments. However, this dataset has three major shortcomings. First, the dataset is very small, comprising approximately 1.5k posts retrieved from the discussion threads of only 10 news articles. Second, the authors did not release sufficient information to reconstruct the threads and allow systems to consider the parent comments. Third, only a single annotator was used for most of the comments, which makes the annotations less reliable. 

Two other datasets, both non English, also include context-aware annotations. 
\newcite{mubarak2017} provided the title of the respective news article to the annotators, but ignored parent comments. This is problematic when new comments change the topic of the discussion and when replies require the previous posts to be judged. \newcite{Pavlopoulos2017a} used professional moderators, who were monitoring entire threads and were thus able to use the context of the thread to judge for the toxicity of the comments. 
However, the plain text of the comments for this dataset is not available, which makes further analysis difficult. Moreover,
crucially for this study, the context of the comments was not released in any form. 

In summary, of the datasets we know of (Table~\ref{tab:datasets}), only two include context 
\cite{Gao2018,mubarak2017}, and 
this context is limited to the title of the news article the comment was about. 
As discussed above, \newcite{Gao2018} include the parent comments in their dataset, but without sufficient information to link the target comments to the parent ones. Hence \textit{no toxicity dataset includes the raw text of both target and parent comments with sufficient links between the two}. This means that toxicity detection methods cannot exploit the conversational context when being trained on existing datasets. 

Using previous comments of a conversation or preceding sentences of a document is not uncommon in text classification and language modeling. 
\newcite{Mikolov2012}, for example, used \lda to encode the preceding sentences and pass the encoded sentence history to an \rnn language model \cite{Blei2003}. Their approach achieved state of the art language modeling results and was used as an alternative solution (e.g., to \lstm{s}) for the problem of vanishing gradients. \newcite{Sordoni2015} experimented with concatenating consecutive utterances (or their representations) before passing them to an \rnn to generate conversational responses. They reported gains up to 11\% in \bleu \cite{Papineni2002}. \newcite{Ren2016} reported significant gains in Twitter sentiment classification, when adding contextual features. 

\section{Experiments}

\begin{table}
    \centering
    \begin{tabular}{|c|c|c|}
        \hline
        {\small \textbf{Dataset Statistics}} & \textbf{\smalldataset} & \textbf{\largedataset} \\\hline
         \#comments (\oc/\ic) & 250 & 10k/10k\\\hline
         avg.\ length (\oc/\ic) & 100 & 161/161 \\\hline
         \#toxic (\goc/\gic) & 11/16 & 59/151 \\\hline
    \end{tabular}
    \caption{Dataset statistics. \smalldataset contains 250 comments. \largedataset contains 10k comments 
    without (\oc) and 10k comments with context (\ic). Average length in characters. \goc is the group of annotators with no access to context, and \gic the group with context. 
    For each comment and group of annotators, the toxicity scores of the annotators were averaged and rounded to the nearest binary decision (toxic, non-toxic) to compute the number of toxic comments (\#toxic).}
    \vspace*{-3mm}
    \label{tab:dataset_stats}
\end{table}

\subsection{Experiments with \smalldataset for \rqone}

To investigate how often context affects the perceived toxicity of posts, we created \smalldataset, a small Context-Aware Toxicity dataset of 250 randomly selected comments from the Wikipedia Talk Pages (Table~\ref{tab:dataset_stats}). We gave these comments to two groups of crowd-workers to judge their toxicity. 
The first group (\gic, Group with Context) was also given access to the parent comment and the discussion title, while the second group (\goc, Group with No context) was provided with no context. No annotator could belong to both groups, to exclude the case of an annotator having seen the context of a post and then being asked to label the same post without its context. We used the Figure~Eight crowd-sourcing platform, which provided us with these mutually exclusive groups of annotators.\footnote{See \url{https://www.figure-eight.com/}. The annotators were high-performing workers from previous jobs. 
The demographics and backgrounds of the crowdworkers are detailed in \citet{Posch2018}.} 
We collected three judgments per comment, per group. All comments were between 10 and 400 characters long. Their depth in their threads was from 2 (direct reply) to 5. 

We used the parent comment and discussion title only, instead of a larger context (e.g., the entire thread), to speed up our machine learning experiments, and also because reading only the previous comment and the discussion title made the manual annotation easier. In preliminary experiments, we observed that including more preceding comments had the side effect of workers tending to ignore the context completely.\footnote{We experimented with providing the \gic annotators with all the parent comments in the discussion. We also experimented with preselection strategies, such as employing the score from a pre-trained toxicity classifier for a stratified selection and using a list of terms related to minority groups.} We addressed this problem by asking the annotators an extra question: ``Was the parent comment less, more, or equally toxic?''

\begin{figure}[t]
\includegraphics[width=.45\textwidth]{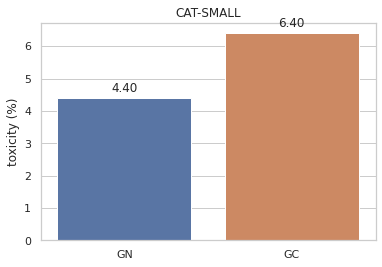}
\vspace*{-4mm}
\caption{Toxicity ratio (\%) of the comments of \smalldataset when using the toxicity labels of \goc (annotators with no context)
or \gic (annotators with context).
The difference is statistically significant ($\textit{P} < .01$).}
\vspace*{-3mm}
\label{fig:ratios_small}
\end{figure}

For each comment and group of annotators, the toxicity scores of the annotators were first averaged and rounded to the nearest binary decision, as in Table~\ref{tab:dataset_stats}. Figure~\ref{fig:ratios_small} shows that the toxicity ratio (toxic comments over total) of \smalldataset is higher when annotators are given context (\gic), compared to when no context is provided (\goc). 
A one-sided Wilcoxon-Mann-Whitney test 
shows this is a statistically significant increase. This is a first indication that providing context to annotators affects their decisions. The toxicity ratio increases by 2 percentage points (4.4\% to 6.4\%) when context is provided, but this is an aggregated result, possibly hiding the true size of the effect of context. The perceived toxicity of some comments may be increasing when context is provided, but for other comments it may be decreasing, and these effects may be partially cancelling each other when measuring the change in toxicity ratio. 

To get a more accurate picture of the effect of context, we measured the number of comments of \smalldataset for which the (averaged and rounded) toxicity label was different between the two groups (\goc, \gic). We found that the toxicity of 4 comments out of 250 (1.6\%) decreased with context, while the toxicity of 9 comments (3.6\%) increased. Hence, perceived toxicity was affected for 13 comments (5.2\% of comments). While the small size of \smalldataset does not allow us to produce accurate estimates of the frequency of posts whose perceived toxicity changes with context, the experiments on \smalldataset indicate that context has a statistically significant effect on the perceived toxicity,
and that context can both amplify or mitigate the perceived toxicity, thus making a first step to addressing our first research question (\rqone). Nevertheless, larger annotated datasets need to be developed to estimate more accurately the frequency of context-sensitive posts in online conversations, and how often context amplifies or mitigates toxicity. 

\subsection{Experiments with \largedataset for \rqtwo}

To investigate whether adding context can benefit toxicity detection classifiers, we could not use \smalldataset, because its 250 comments are too few to effectively train a classifier. 
Thus, we proceeded with the development of a larger dataset. Although the best approach would be to extend \smalldataset, which had two mutually exclusive groups of annotators labeling each comment, we found that the annotation process was very slow in that case, largely because of the small size of annotator groups we had access to in Figure Eight (19 and 23 for \gic and \goc respectively).\footnote{Figure Eight provided us with the two mutually exclusive annotator groups, which could not grow in size.} By contrast, when we did not request mutually exclusive annotator groups, we could get many more workers (196 and 286 for \gic and \goc respectively) and thus annotation became significantly faster.

For this larger dataset, dubbed \largedataset, we annotated 20k randomly selected comments from Wikipedia Talk Pages. 10k comments were annotated by human workers who only had access to the comment in question (group with no context, \goc). The other 10k comments were annotated by providing the annotators also with the parent comment and the title of the discussion (group with context, \gic). 
Each comment was annotated by three workers. We selected comments of length from 10 and 400 characters, with depth in thread from 2 (direct reply) to 5.
Inter-annotator agreement was computed with Krippendorff’s alpha on 123 texts, and it was found to be 0.72\% for \goc and 0.70\% for \gic.

\begin{figure}[t]
\includegraphics[width=.45\textwidth]{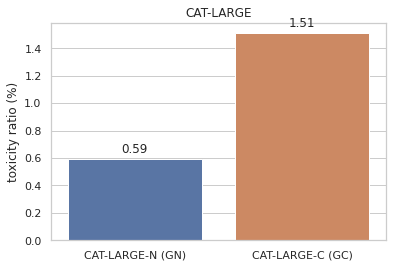}
\vspace*{-4mm}
\caption{Toxicity ratio (\%) of the comments of \largedataset-\oc (10k comments annotated with no context, left) and \largedataset-\ic (10k other comments annotated with context, right). 
For each comment, the toxicity scores of the annotators were first averaged and rounded to the nearest binary decision, as in Table~\ref{tab:dataset_stats}.
The difference is statistically significant
($\textit{P} < .001$).
}
\vspace*{-4mm}
\label{fig:ratios}
\end{figure}
Figure~\ref{fig:ratios} shows that the toxicity ratio increased (from 0.6\% to 1.5\%) when context was given to the annotators. 
A
one-sided Wilcoxon-Mann-Whitney test shows this is a statistically significant 
increase ($P < .001$). Again, the change of toxicity ratio is an indication that context does affect the perceived toxicity, but it does not accurately show how many comments are affected by context, since the perceived toxicity may increase for some comments when context is given, and decrease for others.
Unlike \smalldataset, in \largedataset we cannot count for how many comments the perceived toxicity increased or decreased with context, because the two groups of annotators (\goc, \gic) did not annotate the same comments. 
The toxicity ratios of \largedataset (Fig.~\ref{fig:ratios}) are lower than in \smalldataset (Fig.~\ref{fig:ratios_small}), though they both show a trend of increased toxicity ratio when context is provided. The toxicity ratios of \largedataset are more reliable estimates of toxicity in online conversations, since they are based on a much larger dataset.

We used \largedataset to experiment with both context-insensitive and context-sensitive toxicity classifiers. The former only consider the post being rated (the target comment), whereas the latter also consider the context (parent comment).

\subsection*{Context Insensitive Toxicity Classifiers}

\textbf{\bilstm} Our first context-insensitive classifier is a bidirectional \lstm \cite{Hochreiter1997}. On top of the concatenated last states (from the two directions) of the \bilstm, we add a feed-forward neural network (\ffnn), consisting of a hidden dense layer with 128 neurons and $\tanh$ activations, then a dense layer leading to a single output neuron with a sigmoid that produces the toxicity probability. We fix the bias term of the single output neuron to 
$\log\frac{T}{N}$, where $T$ and $N$ are the numbers of toxic and non-toxic training comments, respectively, 
to counter-bias against the majority (non-toxic) class.\footnote{See an example in \url{http://tiny.cc/m572gz}.} This \bilstm-based model could, of course, be made more complex (e.g., by stacking more \bilstm layers, and including self-attention), but it is used here mainly to measure how much a relatively simple (by today's standards) classifier benefits when a context mechanism is added (see below). 

\smallskip
\noindent\textbf{\bert} At the other end of complexity, our second context-insensitive classifier is \bert \cite{Devlin2018}, fine-tuned on the training subset of each experiment, with a task-specific classifier on top, fed with \bert's top-level embedding of the [\textsc{cls}] token. 
We use \bertbase pre-trained on cased data, with 12 layers and 768 hidden units. We only unfreeze the top three layers during fine-tuning, with a small learning rate (2e-05) to 
avoid catastrophic forgetting. 
The task-specific classifier is the same \ffnn as in the \bilstm classifier.

\smallskip
\noindent\textbf{\bertcctk} We also experimented with a \bert model that is the same as the previous one, but fine-tuned on a sample (first 100k comments) of the \civilcomments dataset (Table~\ref{tab:datasets}). We used the general toxicity labels of that dataset,
and fine-tuned for a single epoch. The only difference of this model, compared to the previous one, is that it is fine-tuned on a much larger training set, which is available, however, only without context (no parent comments). The annotators of the dataset were also not provided with context  (Table~\ref{tab:datasets}). 

\smallskip
\noindent\textbf{\perspective} The third context-insensitive classifier is a 
\cnn-based model for toxicity detection, trained on millions of user comments from online publishers. It is publicly available through the \perspective \textsc{api}.\footnote{\url{https://www.perspectiveapi.com/}} 
The publicly available form of this model cannot be retrained, fine-tuned, or modified to include a context-awareness component. Like \bertcctk, this model uses an external (but now much larger) labeled training set. This training set is not publicly available, it does not include context, and was labeled by annotators who were not provided with context. 

\subsection*{Context Sensitive Toxicity Classifiers}

\begin{figure}
    \centering
    \includegraphics[width=.45\textwidth]{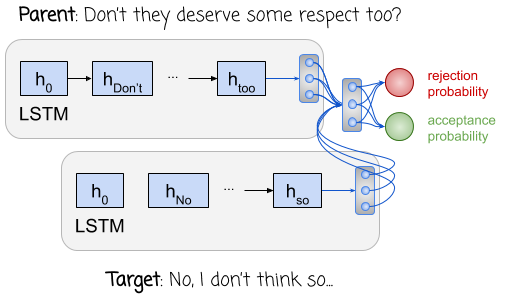}
    \caption{Illustration of \cabilstm. Two \bilstm{s}, shown unidirectional for simplicity, encode the parent and target comment. The concatenation of the vector representations of the two comments is then passed to a \ffnn.}
    \label{fig:carnn}
\end{figure}

\textbf{\cabilstm} In a context-aware extension of the context-insensitive \bilstm classifier, dubbed \cabilstm, we added a second \bilstm to encode the parent comment (Fig.~\ref{fig:carnn}). The vector representations of the two comments (last states from the two directions of both \bilstm{s}) are concatenated and passed to a \ffnn, which is otherwise identical to the \ffnn of the context-insensitive \bilstm. 

\begin{figure}
    \centering
    \includegraphics[width=.45\textwidth]{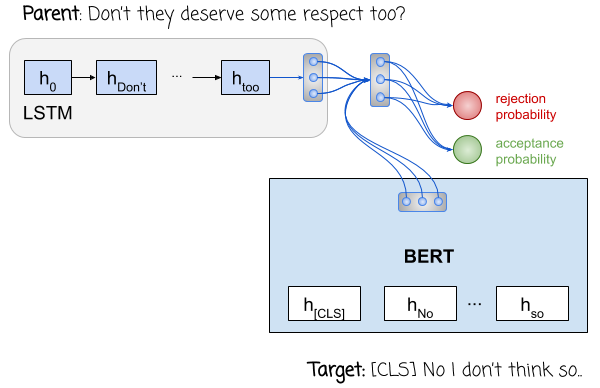}
    \caption{Illustration of \cabert. \bert encodes the target comment. \bilstm (shown unidirectional for simplicity) encodes the parent comment. The vector representations of the two comments are concatenated and passed to a \ffnn.}
    \label{fig:cabert}
\end{figure}

\smallskip
\noindent\textbf{\cabert} We also used a \bilstm to encode the parent in a context-aware extension of the \bert-based classifier, called \cabert (Fig.~\ref{fig:cabert}). Now \bert encodes the target comment, whereas a \bilstm (the same as in \cabilstm) encodes the parent. (We could not use two \bert instances to encode both the parent and the target comment, because the resulting model did not fit in our \textsc{gpu}.) The concatenated representations of the two comments are passed to a \ffnn, which is otherwise the same as as in previous models. \bert is fine-tuned on the training subset, as 
before, and the \bilstm encoder of the parent is jointly trained 
(with a larger learning rate).

\begin{figure}[t]
    \centering
    \includegraphics[width=.35\textwidth,height=5cm]{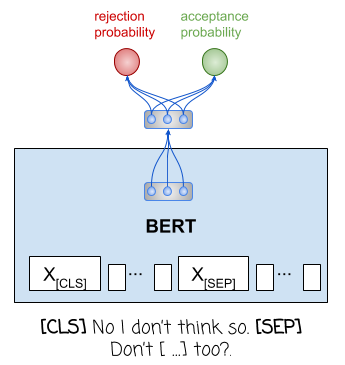}
    \caption{Illustration of \casepbert. A single \bert instance encodes the parent and  target comments, separated by [\textsc{sep}]. The top-level representation of the [\textsc{cls}] token is passed to a \ffnn.}
    \vspace*{-4mm}
    \label{fig:bertsep}
\end{figure}

\smallskip
\noindent\textbf{\casepbert} We also experimented with another context-aware version of the \bert-based classifier, dubbed \casepbert. This model concatenates the text of the parent and target comments, separated by \bert's [\textsc{sep}] token, as in \bert's next sentence prediction pre-training task (Fig.~\ref{fig:bertsep}). Unlike \cabert, it does not use a separate encoder for the parent comment. The model is again fine-tuned on the training subset.

\smallskip
\noindent\textbf{\cabertcctk},\\
\textbf{\caperspective} These are exactly the same as \bertcctk and \perspective, respectively, trained on the same data as before (no context), but at test time they are fed with the concatenation of the text of the parent and target comment, as a naive context-awareness mechanism.

\subsection*{Context Sensitive vs.\ Insensitive Classifiers}

Table~\ref{tab:experimental_results} reports \roc \auc scores, averaged over a 5-fold Monte Carlo (\mc) cross-validation, i.e., using 5 different random training/development/test splits \cite{Gorman2019}; we also report the standard error of mean over the folds. The models are trained on the training subset(s) of \largedataset-\oc (@\oc models) or \largedataset-\ic (@\ic models), i.e., they are trained on comments with gold labels obtained \emph{without} or \emph{with} context shown to the annotators, respectively. All models are always evaluated (in each fold) on the test subset(s) of \largedataset-\ic, i.e., with gold labels obtained \emph{with} context shown to annotators, assuming that those labels are more reliable (the annotators had a broader view of the discussion). In each fold (split) of the \mc cross-validation, the training, development, and test subsets are 60\%, 20\%, and 20\% of the data, respectively, preserving in each subset the toxicity ratio of the entire dataset. We always use the test (and development) subsets of \largedataset-\ic, as always noted. We report \roc \auc, because both datasets are heavily unbalanced, with toxic comments being rare (Fig.~\ref{fig:ratios}). \footnote{Recall that we also fix the bias term of the output neuron of each model (apart from \perspective) to $-\log\frac{T}{N}$, to bias against the majority class. We also tried under-sampling to address class imbalance, but this technique worked best.}

\begin{table}[htb]
    \begin{center}
    \begin{tabular}{|c|c|}
    \hline
    \textbf{model @training} & \textbf{\roc \auc @\ic} \\\hline\hline
    \bilstm@\oc &  56.48$\pm$1.42 \\
    \bilstm@\ic & 56.38$\pm$1.51 \\
    \cabilstm@\oc & 56.13$\pm$1.27 \\
    \cabilstm@\ic & 58.00$\pm$2.70  \\
    \hline
    \bert@\oc & 75.94$\pm$2.73  \\
    \bert@\ic & 73.49$\pm$1.49 \\\
    \cabert@\oc & 74.60 $\pm$3.08  \\
    \cabert@\ic & 74.46$\pm$1.84 \\
    \casepbert@\oc & 73.29$\pm$3.89 \\
    \casepbert@\ic & 73.54$\pm$3.36  \\
    \hline\hline
    \perspective & 79.27$\pm$2.87 \\
    \caperspective & 81.89 $\pm$ 2.79 \\
    \bertcctk & 78.08$\pm$1.50 \\
    \cabertcctk & 81.69$\pm$2.22 \\\hline
    \end{tabular}
    \caption{\roc \auc scores (\%) averaged over five-fold \mc cross-validation (and standard error of mean) for models trained on \largedataset-\oc (@\oc models, gold labels obtained \emph{without} showing context) or \largedataset-\ic (@\ic models, gold labels obtained \emph{with} context). All models evaluated on the test subset of \largedataset-\ic (\auc@\ic, gold labels obtained \emph{with} context). \perspective and \bertcctk were trained on larger external training sets with no context, but are tested on the same test subset (in each fold) as the other models. 
    }
    \vspace*{-3mm}
    \label{tab:experimental_results}
    \end{center}
\end{table}

A first observation from Table~\ref{tab:experimental_results} is that the best results are those of \perspective, \bertcctk, and their context-aware variants (last four rows). This is not surprising, since these systems were trained (fine-tuned in the case of \bertcctk) on much larger toxicity datasets than the other systems (upper two zones of Table~\ref{tab:experimental_results}), and \bertcctk was also pre-trained on even larger corpora. 

What is more surprising is that \emph{any kind of information about the context does not lead to any consistent (or large) improvement in system performance}. \perspective and \bertcctk seem to improve slightly with the naive context-awareness mechanism of concatenating the parent and target text during testing, but the improvement is very small and we did not detect a statistically significant difference.\footnote{We used single-tailed stratified shuffling \cite{Dror2018,Smucker2007}, $P<0.01$, 10,000 repetitions, 50\% swaps in each repetition.}
Training with gold labels obtained from annotators that had access to context (@\ic models) also leads to no consistent (or large) gain, compared to training with gold labels obtained out of context (@\oc models). This is probably due to the fact that context-sensitive comments are few (5.2\% in the experiments on \smalldataset) and, hence, any noise introduced by using gold labels obtained out of context does not significantly affect the performance of the models. 

There was also no consistent (or large) improvement when encoding the parent comments with a \bilstm (\cabilstm, \cabert) or directly as in \bert's next sentence prediction pre-training task (\casepbert). This is again probably a consequence of the fact that context-sensitive comments are few. The small number of context-sensitive comments does not allow the \bilstm- and \bert-based classifiers to learn how to use the context encodings to cope with context-sensitive comments, and failing to cope with context-sensitive comments does not matter much during testing, again since context-sensitive comments are so few.

We conclude for our second research question (\rqtwo) that we found no evidence that context actually improves the performance of toxicity classifiers, having tried both simple (\bilstm) and more powerful classifiers (\bert), having experimented with several methods to make the classifiers context aware, and having also considered the effect of gold labels obtained out of context vs.\ gold labels obtained by showing context to annotators.

\section{Conclusions and Future Work}

We investigated the role of context in detecting toxicity in online comments. We collected and share two datasets for investigating our research questions around the effect of context on the annotation of toxic comments (\rqone) and its detection by automated systems (\rqtwo). We showed that 
context does have a statistically significant effect on toxicity annotation, but this effect is seen in only a narrow slice ($5.2\%$) of the (first) dataset. We also found no evidence that context actually improves the performance of toxicity classifiers, having tried both simple and more powerful classifiers, having experimented with several methods to make the classifiers context aware, and having also considered the effect of gold labels obtained out of context vs.\ gold labels obtained by showing context to the annotators. The lack of improvement in system performance seems to be related to the fact that context-sensitive comments are infrequent, at least in the 
data we collected. 

A limitation of our work is that we considered a narrow contextual context, comprising only the previous comment and the discussion title.\footnote{The discussion title was used only by the human annotators that examined context, not by context-aware systems, which considered only the parent comment.} 
It would be interesting to investigate in future work 
ways to improve the annotation quality when more 
comments in the discussion thread are provided, and also 
if our findings hold when broader context is considered (e.g., all previous comments in the thread, or the topic of the thread as represented by a topic model).
Another limitation of our work is that we used randomly sampled comments. The effect of context may be more significant in conversations about particular topics, or for particular conversational tones (e.g. sarcasm), or when they reference communities that are frequently the target of online abuse. Our experiments and datasets provide an initial foundation to investigate these important directions.

\section*{Acknowledgments}
We thank the anonymous reviewers for their comments. This research was funded in part by Google.

\bibliography{anthology,acl2020}
\bibliographystyle{acl_natbib}

\end{document}